\documentclass[11pt]{article}
\usepackage{colortbl}
 \usepackage{amsmath}
 \usepackage{soul} % Add this in the preamble

\usepackage{float}
\usepackage{tabularx}
\usepackage{amssymb}
 \usepackage{fontawesome5}
\usepackage{booktabs, multirow}
\usepackage{times}
\usepackage{latexsym}
\usepackage{tcolorbox}
\usepackage{booktabs, array, caption, makecell}
\usepackage{pifont}
\usepackage{booktabs}  % For professional tables
\usepackage{graphicx}  % For symbols
\usepackage{colortbl}  % For row colors
\usepackage{amssymb}   % For checkmarks and crosses
\usepackage{multirow}  % For multirow cells

\usepackage[T1]{fontenc}
\usepackage[utf8]{inputenc}
\usepackage{pifont} % For checkmark and xmark symbols
\usepackage{microtype}
\usepackage{booktabs}
\usepackage{multirow}

\usepackage{inconsolata}

\usepackage[preprint]{acl}

\usepackage{times}
\usepackage{latexsym}
\usepackage{authblk}
\usepackage[T1]{fontenc}
\usepackage[utf8]{inputenc}

\usepackage{microtype}

\usepackage{inconsolata}

\usepackage{graphicx}

\title{AutoMedPrompt: A New Framework for Optimizing LLM Medical Prompts Using Textual Gradients}

\author[1,2]{Sean Wu}
\author[4]{Michael Koo}
\author[1,2]{Fabien Scalzo}
\author[2,3]{Ira Kurtz\thanks{Corresponding author: \texttt{IKurtz@mednet.ucla.edu}}}

\affil[1]{Keck Data Science Institute, Pepperdine University}
\affil[2]{Department of Medicine,  University of California, Los Angeles

}
\affil[3]{Brain Research Institute, University of California, Los Angeles
}
\affil[4]{Biological Sciences Division, University of Chicago}

\begin{document}
\maketitle
\begin{abstract}
Large language models (LLMs) have demonstrated increasingly sophisticated performance in medical and other fields of knowledge. Traditional methods of creating specialist LLMs require extensive fine-tuning and training of models on large datasets. Recently, prompt engineering, instead of fine-tuning, has shown potential to boost the performance of general foundation models. However, prompting methods such as chain-of-thought (CoT) may not be suitable for all subspecialty, and k-shot approaches may introduce irrelevant tokens into the context space. We present AutoMedPrompt, which explores the use of textual gradients to elicit medically relevant reasoning through system prompt optimization. AutoMedPrompt leverages TextGrad's automatic differentiation via text to improve the ability of general foundation LLMs. We evaluated AutoMedPrompt on Llama 3, an open-source LLM, using several QA benchmarks, including MedQA, PubMedQA, and the nephrology subspecialty-specific NephSAP. Our results show that prompting with textual gradients outperforms previous methods on open-source LLMs and surpasses proprietary models such as GPT-4, Claude 3 Opus, and Med-PaLM 2. AutoMedPrompt sets a new state-of-the-art (SOTA) performance on PubMedQA with an accuracy of 82.6\%, while also outperforming previous prompting strategies on open-sourced models for MedQA (77.7\%) and NephSAP (63.8\%).
\end{abstract}

\section{Introduction}
In recent years, the widespread adoption of large language models (LLMs) in healthcare has sparked significant interest in their potential to transform medical practice, medical research, and teaching \cite{thirunavukarasu2023large}. However, although proprietary models such as GPT-4 \cite{openai2023gpt4} and Med-PaLM 2 \cite{singhal2022medpalm} have achieved SOTA performance in benchmarks like medical examinations, their high resource consumption in both training and inference create a high barrier of entry for widespread adoption.

As a potential strategy for widespread integration of LLM models in the healthcare field, open-source models provide a valid alternative to proprietary systems because they offer greater flexibility, transparency, and compliance benefits, which are crucial advantages. Moreover, models such as DeepSeek-v3 \cite{liu2024deepseek}, Falcon \cite{almazrouei2023falcon}, Mixtral 8x7B \cite{jiang2024mixtral}, and Qwen2 \cite{chu2024qwen2} have demonstrated competitive performances while using significantly fewer parameters and resources to train than their proprietary counterparts. 

However, in both proprietary and open-source foundation models, the traditional optimization approaches such as fine-tuning and manual prompt engineering, present significant challenges when adapted to the medical field \cite{guluzade2025elmtex}. Fine-tuning methods like low-rank adaption \cite{hu2021lora} require substantial computational resources that are often unavailable to healthcare institutions, and manual prompt engineering demands high domain expertise in both medical knowledge and computational systems.

Automatic prompt optimization frameworks are a more structured approach to enhance performance of out-out-the-box LLMs. Such methods systematically refine prompts to elicit more accurate and contextually relevant responses from LLMs without modifying their underlying weights. Some methods leverage computational techniques like Bayesian search optimization \cite{opsahl2024optimizing}, in-context learning \cite{dong2022survey}, and reinforcement learning \cite{ernst2024introduction}. To address the high barrier to adaptation of open-source LLMs for medical applications, there has been increasing interest in optimizing open-source models for medical question-answering (QA) without explicit parameter updates.\raggedbottom

\subsection{Related Works}
Recent research has introduced automated methods to improve LLMs, which reduces the need for expert intervention and even surpassing parameter updated based methods \cite{pryzant2023automatic, li2024promptist, kepel2024autonomous}.

Reflexion introduced a reinforcement learning framework that enables language agents to self-improve through verbal feedback \cite{shinn2023reflexion}. Instead of fine-tuning model weights, Reflexion agents iteratively generate linguistic reflections on past errors, storing insights in episodic memory for future decision-making. This framework allows LLMs to learn from mistakes without explicit retraining. Reflexion demonstrated significant improvements in coding tasks, decision-making scenarios, and reasoning-based QA. 

Medprompt is another method that enhances the performance of GPT-4 on challenging medical problems through systematic prompt creation. \cite{nori2023generalist}. Medprompt's strategy relies on a composition of multiple general-purpose prompting strategies, including in-context learning and CoT prompting. The approach eliminates the need for expert-crafted exemplars by having the model semantically selecting in-context learning examples based on word embeddings. MedPrompt was the first to achieve an accuracy of greater than 90\% on the MedQA benchmark \cite{nori2023generalist}.

DSPy created a declarative programming model for optimizing LLM pipelines \cite{khattab2023dspy}. DSPy treats LLM calls as modular components within a text transformation graph, where declarative modules replace manually crafted prompt templates. DSPy's framework includes a compiler that iteratively refines prompts and fine-tuning strategies to optimize model performance. This framework achieves substantial performance gains in multi-hop retrieval and complex QA tasks, outperforming expert-crafted few-shot demonstrations in comparison. The approach generalizes well across different tasks and LLM architectures, enabling a scalable and self-improving prompt engineering method.

\begin{figure*}[t] % The '*' makes it span both columns
    \centering
    \includegraphics[width=1.0\linewidth]{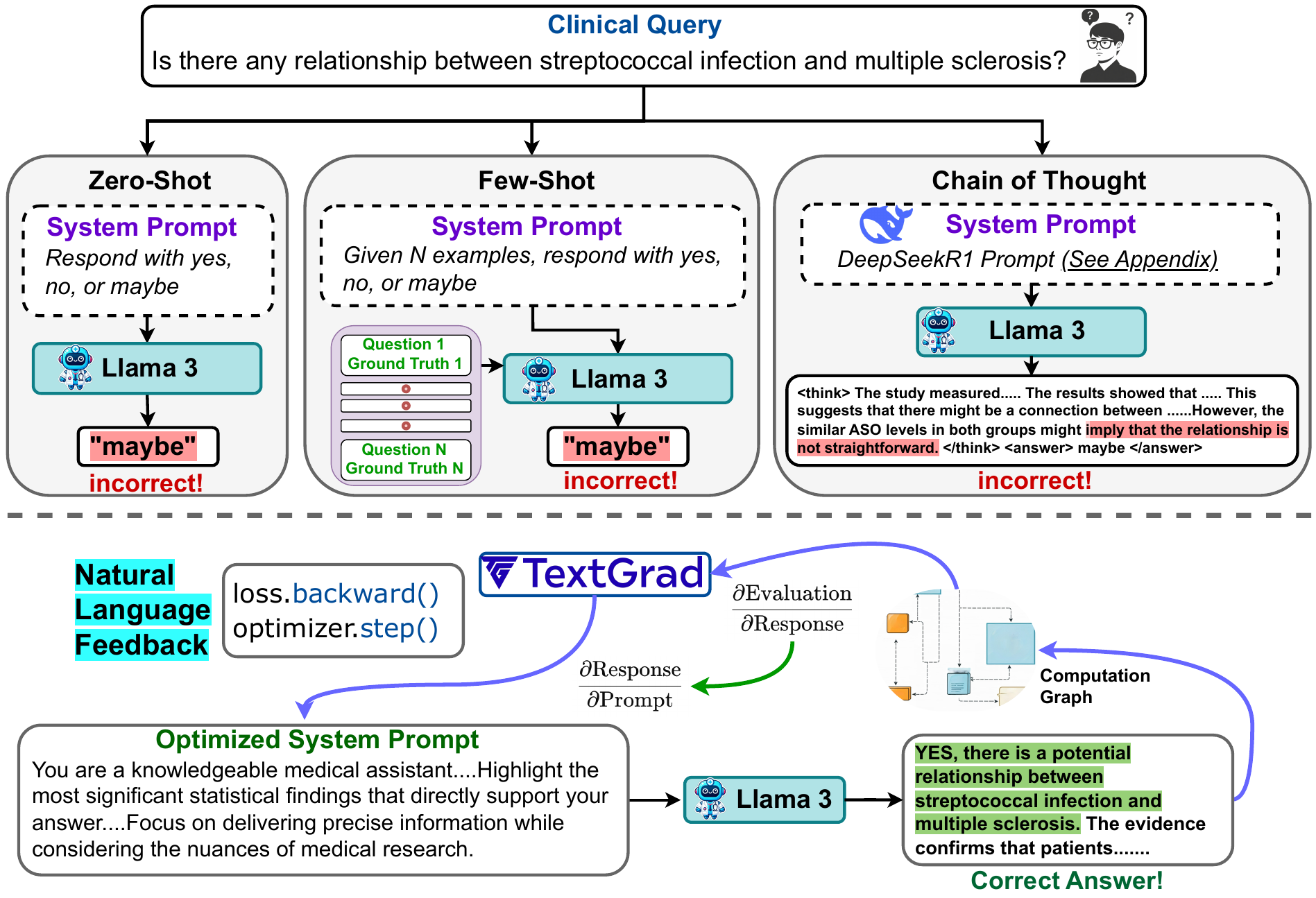} % Adjust width as needed
    \caption{Schematic of AutoMedPrompt, where textual gradients can be superior to traditional prompting strategies.}
    \label{fig:example}
\end{figure*}
OpenMedLM is the current favored SOTA prompting strategy for open-source models, which applies the techniques used by MedPrompt to open-weight LLMs \cite{maharjan2024openmedlm}. OpenMedLM evaluated various open-source models, including Yi34B \cite{01ai2024yi}, on prominent medical benchmarks such as MedQA, MedMCQA \cite{pal2022medmcqa}, and PubMedQA, without resorting to external data or fine-tuning. By utilizing a combination of zero-shot, few-shot, CoT reasoning, and ensemble/self-consistency voting strategies to optimize responses, OpenMedLM demonstrates that strategic prompting alone can outperform fine-tuned specialized models.

Recently, TextGrad is a package that applies the concept of backpropagation with textual gradients for optimizing complex multi-step LLM agents \cite{yuksekgonul2024textgrad}. Specifically, TextGrad's framework constructs a computation graph where nodes such as prompts, tool calls, and responses receive textual feedback from LLMs, functioning as ``textual gradients" that inform iterative refinements. The framework allows for automatic prompt optimization without human intervention, which significantly improves LLM performance on diverse tasks, including question-answering, reasoning, and biomedical applications. TextGrad demonstrates that prompting can be systematically optimized through backpropagation, analogous to differentiable optimization in neural networks.
\subsection{Overview}

We present AutoMedPrompt, a new framework for automatically optimizing medical system prompts with textual gradients. This method outperforms both traditional fine-tuning approaches and black-box proprietary models. Our framework builds upon collective advancements in text-based reflection \cite{shinn2023reflexion}, textual gradients \cite{yuksekgonul2024textgrad}, and transforming generalist LLMs into specialized models through prompting\cite{nori2023generalist}. We demonstrate AutoMedPrompt's effectiveness with Llama-3 \cite{touvron2024llama3} by improving its performance beyond proprietary LLMs and previous prompting strategies across multiple standard medical benchmarks, including MedQA, PubMedQA, and the specialized NephSAP multiple choice question benchmark \cite{wu2024benchmarking}.

AutoMedPrompt achieves SOTA performance on medical benchmarks while significantly reducing optimization constraints. By optimizing only one parameter (the system prompt), AutoMedPrompt helps overcome the computational and expertise barriers that can hinder the widespread adoption of LLMs in the healthcare field. AutoMedPrompt represents a significant advancement in tuning the behavior of general foundation models, enabling them to surpass fine-tuned LLMs through prompt engineering alone. It represents a novel method designed to democratize high-performing medical LLMs using textual gradients, as opposed to CoT or few-shot examples. All code and data are open-source and available on our \href{https://github.com/SeanWu25/AutoMedPrompt}{Github}.
\section{Methodology}

There has been much work done to demonstrate the effectiveness of prompting and, how few-shot and CoT methods can be used to improve the ability to answer medical questions \cite{brown2020language, wei2022chain, ge2023few}. AutoMedPrompt aims to unify these methods by leveraging textual gradients to dynamically optimize prompts for specific tasks rather than a general one-method-fits-all approach.

\subsection{Baseline Prompting Methods}
We evaluated AutoMedPrompt on the following traditional prompting strategies as a baseline.\\\\ \textbf{Zero-Shot Question Answering} with large language models (LLMs) can be formulated as a simple inference problem. Given a medical question from our benchmarks \( Q \), the model \( \mathcal{M} \) predicts the answer \( A^* \) by maximizing the probability distribution over possible answers.\\\\ \textbf{Few-shot in-context learning (ICL)} is another baseline to improve medical QA performance. Given a context containing \( k \) randomly selected example question-answer pairs \( C = \{(Q_i, A_i)\}_{i=1}^{k} \}, \) the model \( \mathcal{M} \) generates an answer \( A^* \) for a new question \( Q \), but now, the model \( \mathcal{M} \) conditions its generated response on the retrieved examples.\\\\ \textbf{Chain-of-Thought (CoT)} is another method we used to improve medical QA by having the LLM use intermediate reasoning before generating the final answer. Here, we leverage the SOTA DeepSeek R1's CoT process template, where the model \( \mathcal{M} \) generates a reasoning step \( R \) and final answer \( A^* \) as  

\[
R, A^* = \mathcal{M}(Q, \mathcal{P}_{\text{CoT}})
\]

See Appendix for the full CoT template.

\subsection{TextGrad Based Optimization}
\begin{figure*}
    \begin{tcolorbox}[
        title=\large One Iteration medical Prompt Optimization, 
        colframe=black!50, 
        colback=white, 
        coltitle=black, 
        colbacktitle=gray!30,
        fonttitle=\bfseries, 
        sharp corners, 
        before upper=\medskip,
    ]
    
    \textbf{Original System Prompt:}  
    \texttt{You are a helpful, creative, and smart assistant.}
    
    \smallskip
    \textbf{Question:}  
    \texttt{Traumatic aortic injury: does the anatomy of the aortic arch influence aortic trauma severity?}
    
    \smallskip
    \textbf{\faRobot{} Llama 3 Response:}  
    Based on the study, the answer is yes, the anatomy of the aortic arch does influence the severity of aortic trauma....
    
    \smallskip
    \textbf{Natural Language Loss} (\(\mathcal{L}_\text{NL}\)):  
    The language model response correctly identifies that the anatomy of the aortic arch influences the severity of aortic trauma. However, the response could be improved by explicitly stating that... The response is accurate but could benefit from a more comprehensive explanation.

    \smallskip
    \textbf{System Prompt Gradient} (\(\nabla_{\text{System Prompt}} \)):  
    To optimize a language model for medical yes/no/maybe questions, refine the system prompt as follows: Medical-Specific Framing, Clarity \& Conciseness, Statistical Awareness, Explicitness, Context Adaptability. 

    \smallskip
    \textbf{Updated System Prompt (1 iteration):}  
    \texttt{You are a concise and evidence-based medical assistant. Provide clear, contextually relevant, and statistically informed answers to yes/no/maybe questions, focusing on key findings without unnecessary details.}
    
\end{tcolorbox}
\end{figure*}
We leverage TextGrad’s autograd engine to (1) automatically construct the computational graphs for medical QA datasets and (2) enable the backpropagation of textual gradients derived from a natural language loss function (LLM call). This method ensures that gradient-based optimization directly influences the LLM's system prompt, which allows  it to dynamically adapt to the current batch of medical queries. Since we focus on a simple medical QA setting, the computational graph for each given query is structured as follows

\begin{align}
    \text{Prediction} &= \mathcal{F}_{\text{LLM}}(\text{Query}, \text{Prompt}),\nonumber \\
    \text{Loss} &= \mathcal{F}_{\text{LLM}}(\text{Prediction}, \text{Eval}\nonumber )
\end{align}

where \(\text{Eval}\) represents a medically validated ground truth taken from a benchmark. TextGrad’s autograd engine internally automatically constructs this computational graph, propagating information through the nodes \( \text{Query} \), \( \text{Prompt} \), and \( \text{Response} \). The optimization process focuses only on improving the Prompt, and ensures that the prompt remains flexible to specific medical contexts and reasoning. This is achieved by requiring the prompt to have gradients during the backpropagation process (requires\_grad = True).

The backpropagation mechanism in TextGrad is conceptually similar to training a neural network, where the optimizable parameter is the Prompt. Given the natural language loss function \( \mathcal{L}_\text{NL} \) (LLM call), the TextGrad generates textual gradients to determine how the \( \text{Response} \) should be refined to better align with the ground truth answer \( G \). This corresponds to computing the partial textual derivative

\begin{align}
    \frac{\partial \mathcal{L}_\text{NL}}{\partial \text{Response}} = \nabla_{\text{LLM}} (\text{Response}, G)\nonumber ,
\end{align}

which is really just the LLM-driven textual gradient  \( \nabla_{\text{LLM}} \), which generates structured linguistic feedback such as
\begin{center}
    \begin{tcolorbox}[colframe=black!75, colback=gray!10, sharp corners]
        \textit{"The response should provide a more detailed explanation of differential diagnoses for the given symptoms."}
    \end{tcolorbox}
\end{center}

This textual gradient then propagates through the computational graph to evaluate the effect of the Prompt on the loss function. This provides insight into how the prompt should be modified to improve the medical response accuracy and reasoning.

\begin{align}
    \frac{\partial \mathcal{L}_\text{NL}}{\partial \text{Prompt}} = \nabla_{\text{LLM}} \left( \text{Prompt}, \text{Response}, \frac{\partial \mathcal{L}_\text{NL}}{\partial \text{Response}} \right)\nonumber .
\end{align}

As an example, the LLM can assess how the system prompt contributed to the response and suggests refinements such as

\begin{center}
    \begin{tcolorbox}[colframe=black!75, colback=gray!10, sharp corners]
        \textit{"The prompt can be improved by instructing the model to prioritize evidence-based treatment recommendations."}
    \end{tcolorbox}
\end{center}

By iteratively applying Textual Gradient Descent (TGD), the clincal system prompt is refined using these updates. 

\begin{align}
    \text{Prompt}^{(t+1)} = \text{TGD.Step} \left( \text{Prompt}^{(t)}, \frac{\partial \mathcal{L}_\text{NL}}{\partial \text{Prompt}} \right)\nonumber .
\end{align}

Over multiple batch iterations, this optimization process ensures that the system prompt updates to generate more medically accurate responses, improve alignment with medical knowledge, and enhanced clarity. An example of one step of this optimization process can be found in the table above.

\subsection{Optimization Trajectory}
When training neural network-based models, it is common for the optimization trajectory to reach suboptimal minima \cite{bashir2010information} or even experience mode collapse in GANs \cite{durall2020combating}. The same issue arises when using textual gradients to optimize any part of a TextGrad computation graph. Often, the $\nabla \text{Textual Gradient}$ leads to the optimization of suboptimal prompts that do not improve or even degrade performance on some benchmarks.

Accordingly, if $p_i$ is the system prompt at training iteration $i$, let $\mathcal{A}(p_i)$ represent the validation accuracy computed on a set of medical questions and answers. The optimization process is constrained by only allowing a system prompt update $p_{i+1}$ if it achieves a higher validation accuracy than the previous best prompt $p^*$. Mathematically, this can be expressed as

\[
p_{i+1} = \begin{cases}
p_{\text{new}} & \text{if } \mathcal{A}(p_{\text{new}}) > \mathcal{A}(p^*), \\
p_i & \text{otherwise},
\end{cases}
\]

where $p_{\text{new}}$ is the candidate prompt generated by the optimization step. If the updated prompt does not perform better, the system reverts to the previous prompt $p_i$, and another batch of training is completed. Finally, we define a stopping criterion to terminate the optimization process if no improvements are observed within $n$ iterations. By doing so, we ensure that every update of the system prompt will directly improve the medical question answering ability of LLama 3.
\section{Experiments}
We tested AutoMedPrompt's textual gradient based optimization on several medical benchmarks, including general popular QA benchmarks like MedQA and PubMedQA, as well as the domain-specific NephSAP, which focuses on nephrology multiple-choice questions. For each benchmark, we tested the zero-shot performance of Llama 3 alongside the randomized few-shot, CoT reasoning with a DeepSeek R1 prompt, and finally prompt optimization with textual gradients.

\subsection{Proprietary and Open-Source LLMs}
Our goal is to combine AutoMedPrompt with the open-source LLM LLama 3 to surpass both proprietary foundation models and also fine-tuned specialists models.\\\\ \textbf{Llama 3} is a foundation language model created by Meta AI \cite{dubey2024llama}, which has models ranging from 8 billion, 70 billion, to 405 billion parameters. The context window of Llama 3 is 128k tokens. Llama 3 is multilingual and was shown to be superior at generating code. Llama 3 is fully open-source, and in this study, we used the 70-billion-parameter version because it is the perfect tradeoff with fitting on commonly used GPUs vs. quality of response.\\\\ \textbf{GPT-4} is OpenAI's fourth-generation Generative Pre-trained Transformer, which is a large-scale multimodal model capable of processing both text and images to generate human-like text outputs \cite{openai2023gpt4}. In this study we compare AutoMedPrompt with Llama 3 to both GPT-4 and GPT-4 with MedPrompt.
\\\\\textbf{Claude 3 Opus} is another proprietary LLM we compare that was designed for complex reasoning tasks and is capable of processing both text and images. It surpassed benchmarks in mathematics, programming, and logical reasoning. \\\\ \textbf{Meditron} is an open-source LLM adapted from Meta's Llama-2 \cite{chen2023meditron}. We use the 70-billion-parameter version. After fine-tuning, Meditron-70B outperforms models like GPT-3.5 and Flan-PaLM \cite{chung2024scaling} on various medical reasoning tasks and performs comparably to GPT-4 and Med-PaLM 2. Meditron is trained on custom-curated medical corpora, including PubMed articles and abstracts, which makes it promising for medical applications. We used Ollama's python library to run Meditron 70b.\\\\\textbf{Med-PaLM 2} is the final proprietary LLM we evaluated. It was created by combining the base language model (PaLM 2) with specialized fine-tuning in the medical domain and innovative prompting strategies, including a novel ensemble refinement approach \cite{singhal2025toward}. 

\subsection{Benchmarks}
\textit{MedQA} is a benchmark that consists of 1273 multiple-choice questions based on the US medical licensing exam (USMLE). The dataset covers both English and Chinese, but our evaluations were only on the English testing subset. The questions were taken from the USMLE Step 1, 2, and 3 exams and are not multimodal.

\textit{PubMedQA} is a curated question-answering dataset based on PubMed abstracts. The language models are asked to respond with either “yes”, “no”, or “maybe.” Each question format is as follows: The reasoning-required setting is a short context from a PubMed abstract, and the goal is to leverage the context to answer the question. A training and testing set is provided.

\textit{NephSAP} dataset comprises 858 multiple-choice questions and answers in the subspecialty medical field of nephrology, drawn from the Nephrology Self-Assessment Program (nephSAP). The questions are contextualized by patient scenarios. In a prior study \cite{wu2024benchmarking}, we demonstrated that open-source models did poorly \cite{wu2024adversarial} compared to GPT-4. The poor results of open-source models \cite{wu2024benchmarking} was a motivation for the current study to improve their success.
\subsection{Experimental Settings}
For system prompt optimization, we require a training set, a development set, and a testing set. The MedQA dataset consists of 10.2k training examples and 1,273 testing examples. To create a development set, we randomly sample 50 examples from the training set. MedQA follows a multiple-choice question format with four possible answer choices.

PubMedQA consists of 500 training examples and 500 randomly selected testing examples. A development set is created by randomly selecting 50 examples from the training set. Unlike MedQA, PubMedQA includes a reasoning section where a short snippet from a PubMed abstract is provided as context, and the model must classify the answer as yes, no, or maybe.

The NephSAP dataset contains 858 questions. The split randomly allocates 50 selected questions to the development set, 500 to the testing set, and the remainder to the training set. 

To access Llama 3, we leveraged Together AI's API platform and used TextGrad's BlackBoxLM function for the forward pass. We set the backwards engine of TextGrad for the loss and gradient backpropagation to OpenAI's GPT-4o and GPT-4o-mini.

\subsection{Evaluating LLM-Generated Responses}
To extract and evaluate the responses generated by the LLMs without the usage of an evaluator API, we define a more systematic approach using regular expressions. The goal is quite simple, to identify and extract the LLM selected answer choice \( \{A, B, C, D, E\} \).

For every LLM generated response, we used the simple expression

\begin{equation}
    \mathcal{M} = \text{Regex}(\backslash b[A-E]\backslash b)\nonumber
\end{equation}

where \( \mathcal{M} \) represents the first detected instance of a valid answer choice. This is checked both manually and by system prompt instructions. If a match \( \mathcal{M} \) is found, the extracted character is taken as the predicted answer. This is similar for DeepSeek R1's chain of thought prompting, where the correct answer is explicitly stated within \texttt{<answer>} tags. If this is the case, we use a similar regular expression to extract the answer within the \texttt{<answer>} field. We follow a similar process for the yes/no/maybe extraction for PubMedQA.
\begin{table*}[!htbp]
    \centering
    \small
    \renewcommand{\arraystretch}{1.1}
    \setlength{\tabcolsep}{4pt}  % Adjust column spacing for better fit

    % Define dark green and dark red colors
    \definecolor{darkgreen}{RGB}{0,100,0}
    \definecolor{darkred}{RGB}{139,0,0}

    % Resize table to fit exactly within text width
    \resizebox{\textwidth}{!}{%
    \begin{tabular}{lcccccc}
        \toprule
        \multirow{2}{*}{\textbf{Prompting Strategy}} & \multirow{2}{*}{\textbf{Model Size}} & \multirow{2}{*}{\textbf{Open-Weight}} & \multirow{2}{*}{\textbf{Avoids Finetuning}} & \multicolumn{3}{c}{\textbf{Accuracy (\%)}} \\
        \cmidrule(lr){5-7}
        & & & & \textbf{PubMedQA} & \textbf{MedQA} & \textbf{NephSAP} \\
        \midrule
        \rowcolor[HTML]{EFEFEF} 
        \textit{Heuristic Baselines} & & & & & & \\
        Random Choice & - & -& - & 33.3 & 25.0 & 24.1 \\
        Human Performance & - & -& - & 78.0 & 60.0 & 76.0 \\
        \midrule
        \rowcolor[HTML]{EFEFEF} 
        \textit{Large Language Models} & & & & & & \\
        GPT-4-base & $\sim$1.76T & \textcolor{darkred}{\ding{55}} & \textcolor{darkgreen}{\ding{51}} & 80.4 & 81.7 & 63.6 \\
        Claude 3 Opus &  $\sim$100B & \textcolor{darkred}{\ding{55}} & \textcolor{darkgreen}{\ding{51}} & 74.9 & 64.7 & 40.8 \\
        Med-PaLM 2 & $\sim$1.6T & \textcolor{darkred}{\ding{55}} & \textcolor{darkred}{\ding{55}} & 81.8 & 85.4 & N/A \\
        MEDITRON & 70B & \textcolor{darkgreen}{\ding{51}} & \textcolor{darkred}{\ding{55}} & 80.0 & 65.4 & 29.8 \\
       
        GPT-4 (Medprompt) & $\sim$1.76T & \textcolor{darkred}{\ding{55}} & \textcolor{darkgreen}{\ding{51}} & 82.0 & 90.2 & N/A \\
        OpenMedLM & 34B & \textcolor{darkgreen}{\ding{51}} & \textcolor{darkgreen}{\ding{51}} & 77.3 & 72.6 & N/A \\
         Llama 3 (Zero-Shot) & 70B & \textbf{\textcolor{darkgreen}{\ding{51}}} & \textbf{\textcolor{darkgreen}{\ding{51}}} & 71.6 & 75.2 & 61.6\\
        Llama 3 (Few-Shot) & 70B & \textbf{\textcolor{darkgreen}{\ding{51}}} & \textbf{\textcolor{darkgreen}{\ding{51}}} &76.0 & 75.4 & 58.8\\
        Llama 3 (DeepSeek R1 CoT) & 70B & \textbf{\textcolor{darkgreen}{\ding{51}}} & \textbf{\textcolor{darkgreen}{\ding{51}}} &71.4 & 76.4 & 48.0 \\
        \textbf{Llama 3 (AutoMedPrompt)} & \textbf{70B} & \textbf{\textcolor{darkgreen}{\ding{51}}} & \textbf{\textcolor{darkgreen}{\ding{51}}} & \textbf{82.6} & \textbf{77.7} & \textbf{63.8} \\
        \bottomrule
    \end{tabular}%
    }  % End of resizebox

    \caption{Comparison of various prompting strategies across multiple medical datasets (PubMedQA, MedQA, and NephSAP).}
    \label{tab:prompting_strategies}
\end{table*}

\section{Results}

We compare the performance of AutoMedPrompt with the open-source Llama 3 70B against previous SOTA methods and proprietary LLMs. Additionally, we conduct ablation studies on Llama 3 using various prompting strategies, along with several heuristic baselines.  
\subsection{Heuristic Baselines}
We compare the LLMs against two heuristic baselines. Firstly, the models should perform well above random choice, which is simply \( \frac{1}{\text{number options}} \). In cases where the number of options varies per benchmark, such as in NephSAP, we take the average. Human performance is also an important reference to determine how far the LLMs are from expert-level performance. The PubMedQA human performance score was obtained from a single test taker, who achieved 78\%, which is higher than most LLMs. For MedQA and NephSAP, we use the passing score as the human benchmark performance (see Table 1).

\subsection{AutoMedPrompt Performance}
In the PubMedQA benchmark, AutoMedPrompt outperforms not only proprietary models such as GPT-4 by 2.2\% but also previous prompting strategies like OpenMedLM by 5.3\% and Gpt-4 with Medprompt by 0.6\%. AutoMedPrompt with Llama 3 achieves an accuracy of 82.6\% on PubMedQA, which is 11\% higher than Llama 3's zero-shot performance. This approach also improves accuracy by 6.6\% compared to few-shot prompting. Furthermore, learning the prompt automatically with textual gradients leads to an 11.2\% improvement over the fixed DeepSeek R1 CoT prompt. More comparisons against Med-Palm 2 and MEDITRON can be found in Table 1.

We observe similar results in the MedQA benchmark, where AutoMedPrompt surpasses zero-shot performance by 2.5\%, few-shot performance by 2.3\%, and CoT prompting by 1.3\%. While the improvement is less pronounced than in the PubMedQA benchmark, textual gradients still enhance performance more effectively than traditional methods. Additionally, AutoMedPrompt outperforms the previous SOTA open-source model, OpenMedLM, by 5.1\%.  

Finally, in the NephSAP benchmark, which is a challenging nephrology self-assessment, we observe that traditional prompting strategies not only fail to improve performance, but few-shot prompting decreases accuracy by 2.8\% and CoT prompting reduces performance by 13.6\%. This suggests that more robust prompting techniques may be necessary depending on the nature of the benchmark questions. AutoMedPrompt with Llama 3 outperforms zero-shot performance by 2.2\% and also surpasses GPT-4 by 0.2\%.
\section{Discussion}
AutoMedPrompt formally utilizes textual gradients when automatically optimizing system prompts for medical use cases. By automatically optimizing only one parameter (system prompt), we demonstrate improvements on all three medical benchmarks beyond previous SOTA open-sourced no fine-tuning methods. We also show that we surpassed proprietary models like GPT-4.
\subsection{More Intuitive Prompting}
Backpropagating textual gradients intuitively would be predicted to enhance the robustness of medical LLMs compared to traditional methods as we demonstrated. Few-shot prompting retrieves question-answer pairs relevant to a similar topic or use case. However, when high-quality answers are unavailable, this approach may fail to improve reasoning, especially if the retrieved examples are only loosely related to the given problem. In the worst cases, irrelevant examples can clutter the context window, consuming valuable space without contributing meaningful insights.

CoT prompting strengthens reasoning by explicitly instructing the LLM to engage in structured thinking before reaching a conclusion. However, not all benchmarks and medical tasks benefit from the same CoT strategies. A one-size-fits-all CoT approach may be suboptimal when different tasks require distinct reasoning approaches.

TextGrad optimization enables more precise system prompts tailored to each specific task. Instead of relying on generic CoT strategies, textual gradients dynamically refine prompts to match the reasoning patterns essential for optimal performance.

For instance, lets consider the differences in system prompts optimized for two distinct medical benchmarks.

\begin{table}[h]
    \centering
    \renewcommand{\arraystretch}{1.1}
    \setlength{\tabcolsep}{3pt}
    \captionsetup{justification=centering}
    \begin{tabular}{@{}p{2.0cm} p{5.5cm}@{}}
        \rowcolor{gray!20} % Light gray header row
        \toprule
        \textbf{Benchmark} & \textbf{Optimized Prompt} \\
        \midrule
        \textbf{NephSAP} & \textit{"You are a precise medical assistant...consider differential diagnoses and reference guidelines...ensure clarity and logical reasoning, and \hl{address misconceptions or counterarguments.}"} \\
        \midrule
        \textbf{PubMedQA} & \textit{"You provide clear, concise, evidence-based answers...\hl{encourage further investigation when findings are preliminary} but maintain assertiveness...focus on precision while considering research nuances."} \\
        \bottomrule
    \end{tabular}
    \caption{Optimized prompts for medical QA datasets}
    \label{tab:optimized_prompts}
\end{table}

From these optimized prompts, we can observe that success in NephSAP requires reasoning through differential diagnoses and addressing counterarguments, while success in PubMedQA emphasizes further investigation of research findings while maintaining clarity and assertiveness.

This shows that while CoT prompting is beneficial, a more granular approach guided by textual gradients is optimal for building more robust medical LLMs.
\subsection{Why AutoMedPrompt Beats Fine-Tuning}
The idea of leveraging a generalist foundation model to outperform a fine-tuned specialist model was introduced in MedPrompt \cite{nori2023generalist}, where they used GPT-4 to achieve SOTA results on benchmarks like PubMedQA. It is not surprising that prompting with textual gradients allows for even further improvement compared to semantic few-shot examples and CoT reasoning, because it enables a more precise adaptation of system prompts to each unique medical task. For example, the prompt optimization for a cardiology question-answering scenario would differ from that of ophthalmology. This is where textual gradients become useful. We found that automatic prompt engineering allows open-sourced, out-of-the-box models such as Llama 3 to surpass their proprietary counterparts and outperform previous strategies like CoT prompting and few-shot examples. Finally, this framework is generalizable to domains beyond medicine, where textual gradients provide instructions on how to refine task-specific system prompts.
\section{Conclusion}
In conclusion, we introduce AutoMedPrompt, which is an open-source framework leveraging textual gradients for automatic medical prompt optimization. AutoMedPrompt enhances the open-source model Llama 3, surpassing proprietary models such as GPT-4 and Claude 3 Opus. Additionally, it outperform the previous SOTA OpenMedLM in open-source prompt engineering on the PubMedQA and MedQA benchmarks. The AutoMedPrompt code is fully open-source and accessible via a simple command-line interface. In this paper, we focus on automatic optimization for medical system prompts, but incorporating an additional step for solution refinement via TextGrad could further improve results. Additionally, more work on interpretability is necessary to develop a more theoretical understanding of why specific prompts are capable of enhancing benchmark performance.

\section*{Limitations}

There are a few limitations to this study. Backpropagating textual gradients leads to rapid convergence when updating just one parameter. While we address this issue by using a batch size, it may still not be optimal. This results in underutilization of the training data, as the training converges before all the provided examples are processed. Such premature convergence is suboptimal for achieving the most robust learning. In future work, we plan to explore methods such as active learning and coreset selection to more systematically sample elements of the training set. A similar issue arises with the validation set. With only 50 random samples chosen, it is highly likely that the validation set is not representative of the dataset as a whole. This can lead to terminating the training process before reaching the optimal system prompt parameter. Increasing the size of the validation set is likely necessary to address this limitation. Another limitation is the absence of GPT-4 with MedPrompt and OpenMedLM results using the NephSAP benchmark. Previous SOTA methods, such as these, require extensive preprocessing, including $K$-nearest neighbors, rerunning questions $n$ times, and employing CoT reasoning. However, these approaches are too computationally expensive for our API-based setup. Finally, Med-Palm 2 was not available for this study due to lack of permission, which could have provided a valuable new insight to the work.

\section*{Acknowledgments}
Ira Kurtz is supported in part by funds from the Smidt Family Foundation, Factor Family Foundation,
Davita Allen Nissenson Research Fund, Ralph Block Family Fund and Kleeman Family Fund.

\bibliography{custom}

\appendix
\pagebreak
\section{DeepSeek R1 CoT Template}
\begin{tcolorbox}[colback=gray!10!white, colframe=black, title=Chain-of-Thought (CoT) Conversation Format]
A conversation between User and Assistant. The user asks a question, and the Assistant solves it. The assistant first thinks about the reasoning process in the mind and then provides the user with the answer. The reasoning process and answer are enclosed within \texttt{<think>} \texttt{</think>} and \texttt{<answer>} \texttt{</answer>} tags, respectively, i.e.,

\texttt{<think> reasoning process here </think>} \\
\texttt{<answer> answer here </answer>}
\end{tcolorbox}

% Example Conversation Box
\begin{tcolorbox}[colback=gray!5!white, colframe=black, title=Example Conversation with GPT-4o]

% User Question
\begin{tcolorbox}[colback=gray!15!white, colframe=black, title=User]
Which of the following is an effective treatment for a viral infection? \\
(A) Antibiotics \quad (B) Rest and hydration \quad (C) Painkillers \quad (D) Vaccines
\end{tcolorbox}

% Assistant's Response
\begin{tcolorbox}[colback=gray!20!white, colframe=black, title=Assistant]
<think> Viral infections cannot be treated with antibiotics because they only target bacteria. Painkillers (e.g., ibuprofen) help with symptoms but do not fight the virus. Vaccines are preventive, not a treatment. The best approach is rest and hydration to support the immune system. </think>
<answer> B </answer>
\end{tcolorbox}

\end{tcolorbox}

\label{sec:appendix}
\onecolumn
\section{PubMedQA Additional Example}

% ==========================
% PubMedQA Benchmark
% ==========================
\begin{tcolorbox}[
    colback=white,
    colframe=black,
    sharp corners,
    title=\textbf{PubMedQA: Does early migraine treatment shorten time to headache peak and reduce its severity?},
    fonttitle=\bfseries,
    width=\textwidth,
    enlarge left by=0mm, 
    enlarge right by=0mm
]
\textbf{Zero-Shot:} No

\textbf{Few-Shot:} No

\textbf{CoT:}  
\begin{quote}
<think> Let's break down the question and analyze the information provided. The study aimed to explore whether early treatment would shorten the duration of headache from headache onset to its peak and reduce headache severity at peak. The results show that early intervention was associated with a shorter time from headache onset to peak, but not with reduced headache severity at peak. </think>  
<answer> Maybe </answer>
\end{quote}

\textbf{AutoMedPrompt (Best Response):}  
\begin{quote}
Yes, early migraine treatment shortens the time to headache peak, but no, it does not reduce its severity. The evidence confirms that treating migraines within 15 minutes of onset reduces the time from headache onset to peak, with a mean time of 1.9 hours compared to 8.9 hours for those who waited 4 or more hours. This finding is supported by multivariate analysis, which shows that early treatment is significantly associated with shorter time from onset to headache peak. However, despite shortening the time to peak, early treatment did not reduce headache severity at peak. The use of triptans was also found to be significantly associated with shorter time from onset to headache peak, highlighting the importance of considering medication type in migraine management.
\end{quote}

\textbf{Ground Truth:} Yes
\end{tcolorbox}
\pagebreak
\section{MedQA Additional Example}
% ==========================
% MedQA Benchmark
% ==========================
\begin{tcolorbox}[
    colback=white,
    colframe=black,
    sharp corners,
    title=\textbf{MedQA: A 23-year-old woman with hereditary spherocytosis. What is true about the rate-limiting enzyme in the affected pathway?},
    fonttitle=\bfseries,
    width=\textwidth,
    enlarge left by=0mm, 
    enlarge right by=0mm
]
\textbf{Zero-Shot:}  
\begin{quote}
Based on the symptoms and physical exam findings, the patient's most likely condition is Hereditary Spherocytosis (HS), a genetic disorder affecting the red blood cell membrane. Now, let's analyze the options related to the rate-limiting enzyme of the biochemical pathway affected in HS: In HS, the affected enzyme is glycolysis, specifically the rate-limiting enzyme phosphofructokinase-1 (PFK-1).
\end{quote}

\textbf{Few-Shot:}  
\begin{quote}
Based on the patient's symptoms and physical examination, the most likely diagnosis is hereditary spherocytosis, a genetic disorder affecting the red blood cell membrane.  
The correct answer is: A. It is stimulated by ATP.
\end{quote}

\textbf{CoT:}  
\begin{quote}
<think> The patient's symptoms of fatigue, shortness of breath, and conjunctival pallor suggest anemia. The presence of echinocytes on the peripheral blood smear is also consistent with anemia. The family history of similar issues suggests a possible genetic disorder. Considering the patient's symptoms and the presence of echinocytes, the most likely condition is hereditary spherocytosis, which is a defect in the red blood cell membrane. This defect is often caused by mutations in genes involved in the glycolytic pathway, specifically in the production of ATP. </think>  
<answer> D. It is inhibited by AMP </answer>
\end{quote}

\textbf{AutoMedPrompt (Best Response):}  
\begin{quote}
The correct answer is C. It is inhibited by protein kinase A activity. Based on the patient's symptoms and physical exam findings, the most likely condition is hereditary spherocytosis, a genetic disorder affecting the red blood cell membrane.
\end{quote}

\textbf{Ground Truth:} C
\end{tcolorbox}

\pagebreak
\section{NephSAP Additional Example}
% ==========================
% NephSAP Benchmark
% ==========================
\begin{tcolorbox}[
    colback=white,
    colframe=black,
    sharp corners,
    title=\textbf{NephSAP: Which of the following agents could be used for his osteoporosis?},
    fonttitle=\bfseries,
    width=\textwidth,
    enlarge left by=0mm, 
    enlarge right by=0mm
]
\textbf{Zero-Shot:}  
\begin{quote}
What a complex case! Considering the patient's medical history, particularly his severe COPD, vertebral fracture, and osteoporosis, we need to choose an agent that is effective for osteoporosis treatment while minimizing potential risks and interactions.
\end{quote}

\textbf{Few-Shot:}  
\begin{quote}
Based on the context and the patient's history, I would recommend: D. Romosozumab.
\end{quote}

\textbf{CoT:}  
\begin{quote}
<think> Let's analyze the patient's condition and medical history. He has severe COPD, coronary artery disease, and peripheral vascular disease, which suggests that he may not be a good candidate for medications that could exacerbate these conditions. He has a history of fractures, including a nontraumatic vertebral fracture, indicating osteoporosis. </think>  
<answer> The correct answers are B. Denosumab and D. Romosozumab. </answer>
\end{quote}

\textbf{AutoMedPrompt (Best Response):}  
\begin{quote}
Based on the patient's medical history and laboratory results, I would recommend option C, Teriparatide.
\end{quote}

\textbf{Ground Truth:} C
\end{tcolorbox}

\end{document}